\documentclass{article}

\usepackage[preprint]{neurips_2026}

\usepackage[utf8]{inputenc} % allow utf-8 input
\usepackage[T1]{fontenc}    % use 8-bit T1 fonts
\usepackage{hyperref}       % hyperlinks
\usepackage{url}            % simple URL typesetting
\usepackage{booktabs}       % professional-quality tables
\usepackage{amsfonts}       % blackboard math symbols
\usepackage{nicefrac}       % compact symbols for 1/2, etc.
\usepackage{microtype}      % microtypography
\usepackage{xcolor}         % colors
\usepackage{amsmath}
\usepackage{natbib}
\usepackage{amssymb}
\usepackage{graphicx}
\usepackage{wrapfig}

\newcommand{\bh}{\mathbf{h}}

\newcommand{\bw}{\mathbf{w}}
\newcommand{\br}{\mathbf{r}}
\newcommand{\bff}{\mathbf{f}}
\newcommand{\bB}{\mathbf{B}}
\newcommand{\bR}{\mathbf{R}}
\newcommand{\bF}{\mathbf{F}}
\newcommand{\bW}{\mathbf{W}}
\newcommand{\bU}{\mathbf{U}}
\newcommand{\bG}{\mathbf{G}}
\newcommand{\bM}{\mathbf{M}}
\newcommand{\bV}{\mathbf{V}}

\newcommand{\bz}{\mathbf{z}}
\newcommand{\bu}{\mathbf{u}}
\newcommand{\bv}{\mathbf{v}}
\newcommand{\bT}{\mathbf{T}}

\newcommand{\bq}{\mathbf{q}}

\newcommand{\bmoves}{\mathbf{m}}
\newcommand{\bBoard}{\mathcal{B}}
\newcommand{\btheta}{\boldsymbol{\theta}}
\newcommand{\scurrent}{\textsc{Current}}
\newcommand{\sopponent}{\textsc{Opponent}}
\newcommand{\sempty}{\textsc{Empty}}
\newcommand{\para}[1]{\noindent\textbf{#1}}

\title{Tensor Product Representation Probes Reveal Shared Structure Across Linear Directions}

\author{%
  Andrew Lee \\
  Harvard University \\
  \texttt{andrewlee@g.harvard.edu} \\
  \And
  Fernanda Viégas \\
  Harvard University\\
  Google DeepMind\thanks{Work done at Harvard. Code: \url{https://github.com/ajyl/tpr_othello}} \\
  \And
  Martin Wattenberg \\
  Harvard University\\
  Google DeepMind$^*$ \\
}

\begin{document}

\maketitle

\begin{abstract}
    While researchers are finding concepts represented as linear directions in language models, a bag of linear directions fails to capture relational structure.
    To better understand this dichotomy, we study a model with known linear representations, but trained in a highly structured domain -- the board game Othello.
    While the model's internal board-state representation is linearly decodable, we find additional structure in the form of tensor product representations (TPRs).
    We train TPR probes to recover shared structure amongst the linear probes, yielding a factorization into square-embeddings, color-embeddings, and a binding matrix that composes them to construct the model's board-state representation.
    We find geometric signatures within the weights of our TPR probe that align with the structure of the board, but perhaps more importantly, that the linear probes can be recovered directly from the parameters of our TPR probe.
    Our findings suggest that directional representations may be projections of more structured underlying representations.
%{\let\thefootnote\relax\footnotetext{Code: \url{https://github.com/ajyl/tpr_othello}}}
\end{abstract}

\section{Introduction}
\label{sec:intro}

As Transformers become widely adopted in various domains, researchers are finding numerous forms in which they represent information.
Among a popular line of work is the \emph{linear representation hypothesis}~\citep{park2023linear}, which posits that Transformers represent concepts as linear directions in the model's activation space.
This hypothesis has been supported by various empirical findings, with rank-1 representations for concepts like sentiment~\citep{tigges2023linear}, toxicity~\citep{lee2024mechanistic}, refusal~\citep{arditi2024refusal}, or even properties like user attributes~\citep{chen2024designing}.

%Building on such hypothesis, researchers have developed various techniques to interpret language models:
%sparse autoencoders (SAEs)~\citep{cunningham2023sparse}, reconstruct the model's activations as sparse linear combinations of concepts, while circuit analyses~\citep{marks2024sparse, dunefsky2024transcoders} trace the computational paths between concept representations and model outputs.

While the simplicity of linearity makes it an appealing representation, it fundamentally lacks structure by treating concepts as a bag of linear directions -- which does not reflect how the world is structured~\citep{wattenberg2024relational}.
How do we reconcile this dichotomy?

We study this question by revisiting prior work with known linear representations in a domain with inherent structure -- board games.
Namely, we study OthelloGPT~\citep{li2022emergent}, a Transformer model trained on the board game Othello.

OthelloGPT provided one of the earliest evidence for linear representations in Transformers.
Namely, \cite{li2022emergent} train a Transformer on move transcripts of Othello: given \emph{only} move sequences as training data, the objective is to predict next legal moves that can follow.
Interestingly, the model learns to internally represent the underlying board-state corresponding to the input moves, despite never being told about the existence of a board to begin with.
Follow-up work by \cite{nanda2023emergent} find \emph{linear} representations of the board-state, in which the latent board-state can be linearly decoded from the model's activations when the board is viewed from an egocentric perspective (see Section~\ref{sec:background}).

\begin{figure}
\centering
\includegraphics[width=0.95\linewidth,trim={0, 2.25cm, 0, 2.55cm},clip]{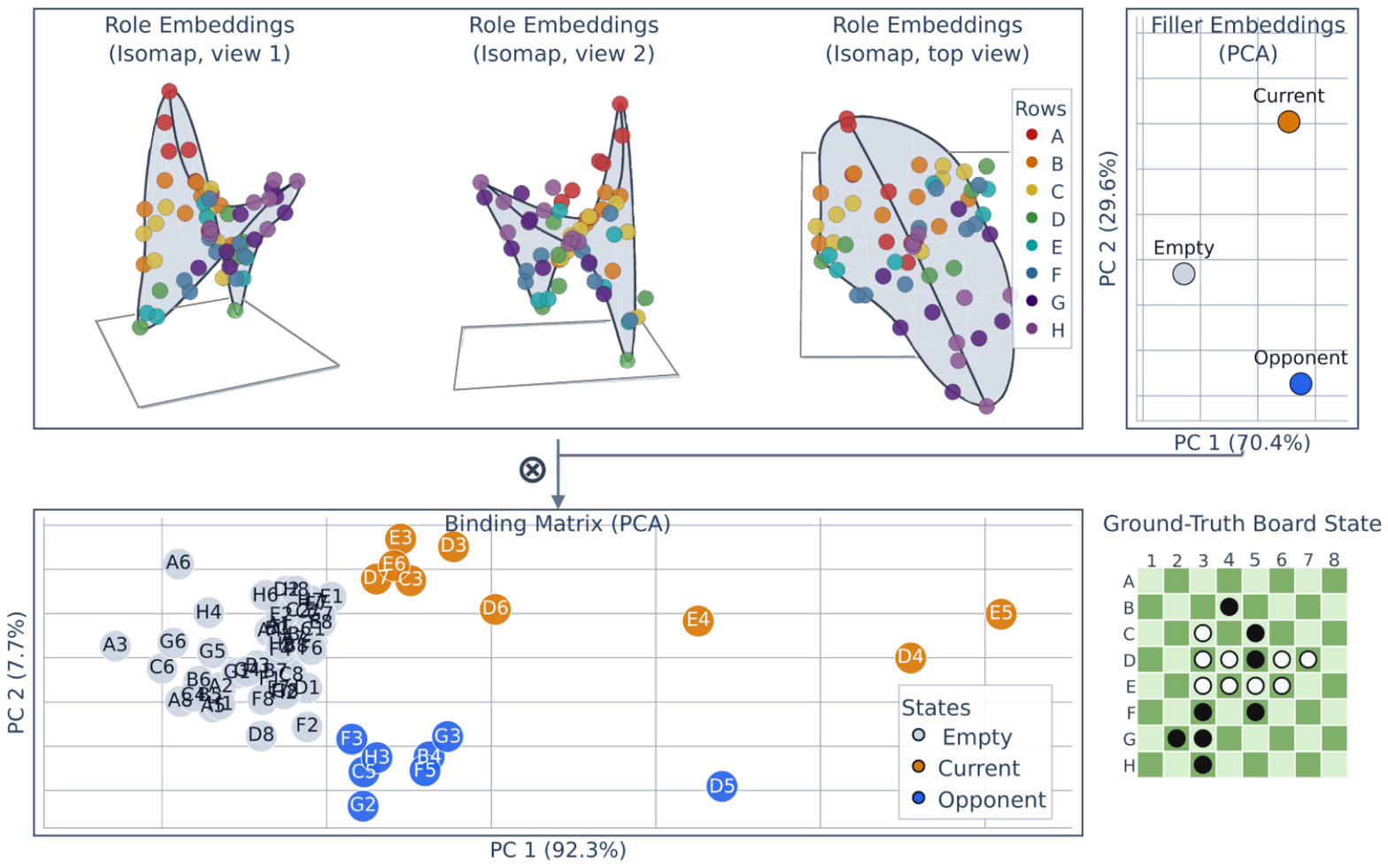}
\caption{\textbf{Tensor product representation probes recover a structured factorization of OthelloGPT's board-state representation}, including
$\mathbf{(i)}$ role (square) embeddings $\bR \in \mathbb{R}^{64 \times d_r}$ (top-left, Isomap), $\mathbf{(ii)}$ filler (color) embeddings $\bF \in \mathbb{R}^{3 \times d_f}$ (top-right, PCA), $\mathbf{(iii)}$ a binding matrix $\bB \in \mathbb{R}^{d_r \times d_f}$ to bind the two objects (bottom-left, PCA) to represent a board-state (bottom-right).
}
\vspace{-5pt}
\label{fig:main_figure}
\end{figure}

However, Othello has additional structure in that squares and color pieces are distinct concepts, and pieces occupy different squares to form a board-state.
Standard linear probing may yield many readout directions, but does not specify whether these directions arise from shared structured factors.
In our work we find that such structure can be recovered with the right decoder architecture.

Namely, \citet{smolensky1990tensor} demonstrates how symbolic structure, such as variable binding, could be represented in connectionist systems, using \emph{tensor product representations} (TPR).
TPR is a general framework for representing complex structures as the binding of simpler components, where the binding operation is implemented as a tensor product.

This framework is a natural fit for Othello, in that the board-state is a binding of two objects: the squares and the colors.
Thus we design and train \textbf{Tensor Product Representation (TPR) probes} to decode OthelloGPT's board-state representations in a factorized manner, allowing us to recover ``square-embeddings'', ``color-embeddings'', and the binding matrix that composes the two objects.

TPR thus factorizes shared structure across linear probes (Figure~\ref{fig:main_figure}).
Furthermore, the learned TPR weights exhibit geometric signatures that align with the structure of the board. 
Lastly, we find that the independently trained linear probes from prior work can be recovered directly from the parameters of our TPR probe.
Put differently, the linear probes are a projection of the more structured TPR probes.

Our findings raise the question of whether linear directions are projections of a richer, compositional underlying structure.

\section{Background, Notations}
\label{sec:background}

In our work we study OthelloGPT~\citep{li2022emergent}, an 8-layer Transformer trained on 20 million game transcripts of the board game Othello.
Given only move sequences as training data ([``D3'', ``E3'', ...]), it predicts with near perfect accuracy the next legal moves that can follow.
Importantly, OthelloGPT has no a priori knowledge of the game nor its rules, or even the existence of an underlying board.

Interestingly, \cite{li2022emergent} demonstrate that OthelloGPT internally represents the underlying board-state in its activations.
Subsequent work~\citep{nanda2023emergent} shows that such board-state representation is \emph{linear} decodable when the board is viewed from an \emph{egocentric} perspective.
Namely, rather than labeling each square of the board as black or white pieces, one must consider whose turn it is to play at each token timestep.
For odd timesteps, because it is black's turn to play, every black piece is labeled $\scurrent$ while white pieces are labeled $\sopponent$, and vice versa on even moves.
%Re-labeling the ground-truth board-state accordingly allows a simple linear decoder to recover the correct board-state representation.
Despite this nuance, we refer to each square's label as ``color'' for simplicity.

While a bag of linear probes can successfully decode the board-state, Othello has additional structure in that squares and colors are distinct concepts.
In our work we demonstrate that such structure can be captured via tensor product representations~\citep{smolensky1990tensor}.

\para{Notations.}
We use boldface lowercase for vectors ($\bh, \bw$), boldface uppercase for matrices ($\bW$), non-boldface for scalars or discrete symbols (e.g., $s, c$), and calligraphics to denote sets or lists ($\mathcal{C}$).

We denote OthelloGPT as $\btheta$, and an input sequence of moves as $\mathcal{X} := [x_0, \dots, x_{T-1}]$ where $x_i$ is a move token (e.g., $B5$).
Every partial move sequence $\mathcal{X}_{:t}$ has a corresponding 8-by-8 board-state, denoted as $\bBoard(\mathcal{X}_{:t}) \in \mathcal{C}^{8 \times 8}$, where $\mathcal{C} := \{\scurrent, \sopponent, \sempty\}$ is the set of possible labels for each square.
For simplicity we refer to square labels $\mathcal{C}$ as ``colors''.
Finally, $s \in \mathcal{S} := \{1,\dots,64\}$ denotes a square index.
\section{Linear Probes vs. Tensor Product Representations}
\label{sec:linear_vs_tpr}

Here we describe linear probes and tensor product representation probes.

\subsection{Probing methods}
\label{subsec:probing_methods}

\para{Linear Probes.}
Assume an input sequence $\mathcal{X}_{:t}$, its corresponding board-state $\bBoard(\mathcal{X}_{:t}) \in \mathcal{C}^{8 \times 8}$, and the model's activations at layer $l$ as $\bh_t^{l} \in \mathbb{R}^d$.
Per square $s$, we train a linear probe $\bW_{s} \in \mathbb{R}^{3 \times d}$ to predict whether the model encodes that square $s$ is occupied by color $c$:
\begin{align}
    \boldsymbol{\ell}_{s}^{probe} = \bW_{s}\bh^l_t \in \mathbb{R}^3, \quad \mathcal{L}_{s}^{probe} = \mathrm{CrossEntropy}(\boldsymbol{\ell}_{s}^{probe}, \bBoard(\mathcal{X}_{:t})_s).
\end{align}
where $\boldsymbol\ell_{s}^{probe}$ is the probe's logits for the three possible colors occupying square $s$ and $\bBoard(\mathcal{X}_{:t})_s \in \mathcal{C}$ is the groundtruth color of square $s$.
Each row of $\bW_{s}$ is a linear readout of whether $s$ is occupied by color $c$.
This results in $64 \times 3 = 192$ separate linear directions to represent the board.

\para{Tensor Product Representation.}
Tensor product representations (TPRs, \citep{smolensky1990tensor}) provide a classical account of how distributed vector representations can encode symbolic structure through role-filler binding.
TPRs represent structured objects by superimposing the outer products of \emph{roles} (e.g., squares of the board) and \emph{fillers} (e.g., colors).

Formally, assume $k$ possible roles and let $\br_i \in \mathbb{R}^{d_r}$ denote a vector representation for role $i$.
Let $y(i)$ denote the filler occupying that role, and $\bff_{y(i)} \in \mathbb{R}^{d_f}$ denote the vector representation of such filler.
Then the resulting binding is
\vspace{-1em}
\begin{align}
    \bB := \sum_{i=1}^k \br_i \otimes \bff_{y(i)} \in \mathbb{R}^{d_r \times d_f}.
\end{align}

This framework fits naturally with Othello board-states.
In our finite-dimensional setting, we identify the tensor product $\br \otimes \bff$ in coordinates as the rank-1 outer product matrix $\br\bff^\top \in \mathbb{R}^{d_r \times d_f}$.
Treating each of the 64 squares as roles and colors as fillers, the board-state can be represented as
\begin{align}
    \bB = \sum_{s\in\mathcal{S}} \br_{s} \otimes \bff_{y(s)} = \sum_{s\in\mathcal{S}}\br_s\bff^\top_{y(s)} \in \mathbb{R}^{d_r \times d_f},
\end{align}
where $\br_{s} \in \mathbb{R}^{d_r}$ is the role vector for square $s$,
$y(s) \in \mathcal{C}$ is the square's corresponding color, and $\bff_{y(s)} \in \mathbb{R}^{d_f}$ is the filler vector for such color.

\para{Training TPR Probes.}
We train TPR probes to decode OthelloGPT's board representation.
The probe learns three sets of weights: role embeddings 
$\bR := [\br_1^\top; \dots; \br_{64}^\top] \in \mathbb{R}^{64 \times d_r}$,
filler embeddings $\bF := [\bff_{empty}^\top; \bff_{current}^\top; \bff_{opponent}^\top] \in \mathbb{R}^{3 \times d_f}$,
and a linear map $\bM \in \mathbb{R}^{d_r \times d_f \times d_{\textrm{model}}}$ from hidden states to a latent binding matrix $\bB$.
Given a hidden state $\bh^l_t$, the probe first constructs the binding matrix $\bB$.

\vspace{-1em}
\begin{align}
    \bB := \bM (\bh^l_t) \in \mathbb{R}^{d_r \times d_f}.
\end{align}
It then scores each square--color pair with a bilinear unbinding:
\begin{align}
    \ell_{s,c} = \br_{s}^{\top} \bB\,\bff_c
\end{align}

The binding matrix $\bB$ encodes which roles (squares) are occupied by which fillers (colors), and unbinding with the learned role $\br_s$ and filler $\bff_c$ recovers the correct color $c$ for square $s$.

\para{Trilinear TPR Probes.}
Note that bindings do not have to be bilinear.
While a natural decomposition for Othello is between the board and colors, the board is also structured with 8 rows and columns.
Thus an alternative \emph{trilinear} TPR probe may encode the state of square $s$ (row $i$, column $j$) as

\vspace{-12pt}
\begin{align}
    \bT := \bM(\bh^l_t) \in \mathbb{R}^{d_{u} \times d_{v} \times d_f}, \qquad \ell_{ij,c} = \langle 
    \bT, \bu_i \otimes \bv_j \otimes \bff_c\rangle
\end{align}

where $\bM \in \mathbb{R}^{d_u \times d_v \times d_f \times d_{model}}$ and $\otimes$ denotes outer products.
The probe learns four sets of weights: \textbf{(i)} a mapping $\bM$ from hidden states to the binding space, \textbf{(ii)} row embeddings $\bU := [\bu_1^\top; \dots;\bu_8^\top] \in \mathbb{R}^{8 \times d_{u}}$, where each row ($\bu_i$) corresponds to a row embedding,
\textbf{(iii)} a column embedding matrix $\bV := [\bv_1^\top; \dots, ;\bv_8^\top] \in \mathbb{R}^{8 \times d_{v}}$,
and \textbf{(iv)} the same color embedding matrix $\bF \in \mathbb{R}^{3 \times d_f}$ as before.

\begin{figure}
\centering
\includegraphics[width=0.88\linewidth]{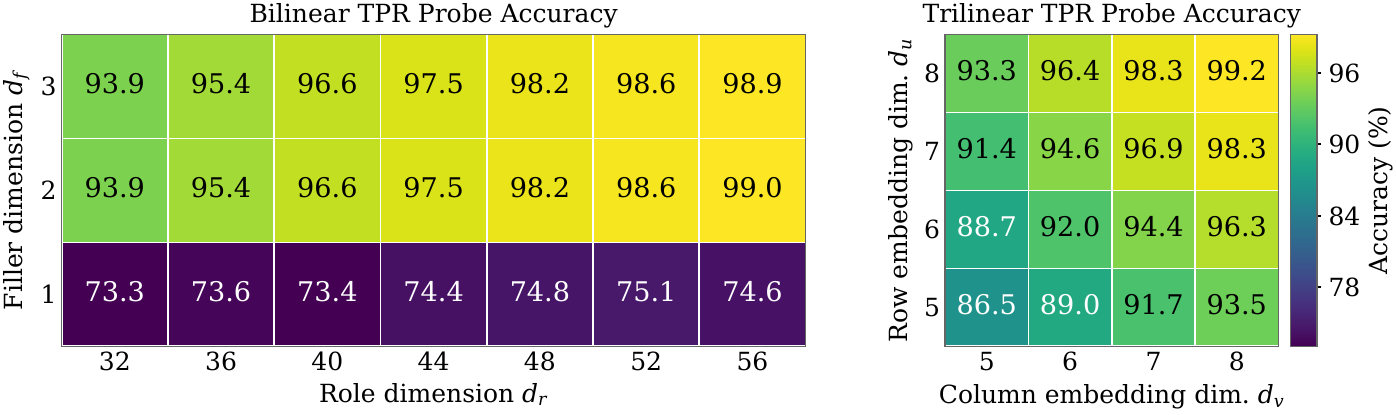}
\caption{\textbf{TPR probe accuracy:}
The board-state can be reconstructed using low-rank role embeddings (rank-$d_r$, or $d_u, d_v$) and filler embeddings (rank-$d_f$).
}
    \label{fig:probe_accuracy}
\end{figure}

\subsection{Probing Results}
\label{subsec:probing_results}

In line with ~\cite{nanda2023emergent}, our linear probes achieve 99\% accuracy.
For TPR probes, Figure~\ref{fig:probe_accuracy} reports the average accuracy from layer 7 (out of 8) over all 64 squares (see Appendix~\ref{appx:additional_results} for all other layers). 
In the case of bilinear, we sweep over multiple values of the role dimension $d_r$ and filler dimension $d_f$.
For trilinear probes we fix $d_f=2$ and sweep over multiple values of $d_u, d_v$.
Note that the TPR probe only requires low-rank role and filler embeddings to achieve ~99\% accuracy (note that $d_{\textrm{model}} = 512$).
Also note that the TPR probe requires at least 2 dimensions for filler embeddings, indicating that high accuracy is not being achieved by role embeddings alone.

It is worth noting that the TPR probes have \emph{fewer} parameters than the linear probe.
The linear probe has $192 \times d_\text{model}$ parameters ($98,304$ with $d_\text{model} = 512$), while bilinear TPRs have $(64 \times d_r) + (3 \times d_f) + (d_r \times d_f \times d_\text{model})$ ($56,582$ with $d_r = 52$, $d_f = 2$; 57.5\% of the linear probe's parameters).
For trilinear probes we have $(8 \times d_u) + (8 \times d_v) + (3 \times d_f) + (d_u \times d_v \times d_f \times d_\text{model})$ ($65,670$ with $d_u=d_v=8$, $d_f=2$).
Despite having fewer parameters and additional structural constraints, the TPR probes achieve near perfect accuracy.

\subsection{Visualizing TPR Probes}
\label{subsec:visualizing_tpr_probes}

Our TPR probe provides a structured decomposition of the model's hidden states into role (square) embeddings, filler (color) embeddings, and a binding matrix that wires the two together.
Figure~\ref{fig:main_figure} visualizes all three components using Isomap and PCA.
The top left panel visualizes the role (square) embeddings $\bR \in \mathbb{R}^{64 \times d_r}$ using Isomap, with points color-coded by each square's row.
The resulting 3D embedding demonstrates a saddle-like manifold with visible features of the board: the upper curve captures rows A through H, while the lower curve captures columns 1 through 8.

The top right panel visualizes the filler (color) embeddings $\bF \in \mathbb{R}^{3 \times d_f}$ using PCA.
Note that principal component (PC) 1 separates $\sempty$ vs. occupied, while PC 2 separates $\scurrent$ vs. $\sopponent$.

These two components, $\bR$ and $\bF$, are based on the TPR probe's parameters alone, and not based on any input.
The bottom left panel visualizes an example of the binding matrix $\bB \in \mathbb{R}^{d_r \times d_f}$, which is input-specific, using PCA.
The bottom right panel shows its corresponding ground-truth board-state.
Note that the binding matrix captures all the correct square--color bindings, while all 64 squares form clusters that mirror the structure of the filler embeddings.
The center four squares (D4, D5, E4, E5) are outliers because according to the rules of Othello, they can never be empty -- the game starts with the 4 squares already filled, and a square that is filled can never turn empty.

\section{Recovered Structure of TPR Probes}
\label{sec:tpr_structure}

Here we study the structure recovered by TPR probes.

\subsection{Interventions}
\label{subsec:interventions}

To validate that the TPR probes not only recover a valid structure but also one that captures causal mechanisms for next-move prediction, we run causal interventions.
We intervene on the model's internal board-state representation and check whether this leads to the expected changes in the model's next-move predictions.
Our intervention methods for both linear and TPR probes are described below.

\para{Setup.}
Recall that $\btheta$ denotes OthelloGPT, $\mathcal{X}$ denotes an input move sequence, and $\bBoard(\mathcal{X}) \in \{\scurrent, \sopponent, \sempty\}^{8 \times 8}$ denotes the corresponding groundtruth board-state.
Let $\boldsymbol{\ell} = \btheta(\mathcal{X}) \in \mathbb{R}^{64}$ be OthelloGPT's logits for next-move predictions, which we reshape and convert to binary predictions using a probability threshold: $\bmoves_{orig} := \mathbb{I}(\text{Softmax}(\boldsymbol{\ell}) > 0.01) \in \{0, 1\}^{8 \times 8}$.

We manipulate the Transformer's internal board-state representation to check whether it is causally responsible for the model's next move predictions.
To do so, let $\bBoard^{(target)}$ be a \emph{target} board-state that we wish the model to represent instead.
To create $\bBoard^{(target)}$, we randomly select a non-empty cell $(i, j)$ from $\bBoard$ and either flip its value (e.g., $\bBoard^{(target)}_{ij} = \scurrent \rightarrow \sopponent$) or set it to empty ($\bBoard^{(target)}_{ij} = \sempty$).
Given the modified board-state $\bBoard^{(target)}$, let $\bmoves_{target} \in \{0, 1\}^{8 \times 8}$ denote its corresponding set of valid next moves according to the rules of Othello.
We validate that every target board-state $\bBoard^{(target)}$ is a legal board-state (i.e., there can be no disconnected ``islands'' of pieces) and that the new set of legal moves $\bmoves_{target}$ does not equal the original set of legal moves $\bmoves_{orig}$.

We then intervene on the Transformer's internal board-state to match $\bBoard^{(target)}$ (which we describe how below), and check how the model's new next-move predictions $\bmoves_{interv}$ compare to $\bmoves_{target}$.

\para{Linear probe interventions.}
Let $\bW_s \in \mathbb{R}^{3 \times d}$ denote the linear probe for square $s$, and $\bw_{s, c} \in \mathbb{R}^d$ denote the row corresponding to color $c$ in $\bW_s$.
Given the hidden-state $\bh$, a high dot-product $\bw_{s, c}^\top \bh$ indicates that square $s$ is occupied by color $c$.
A simple intervention is to add $\bw_{s, c}$ to the hidden state $\bh$:
$\widehat{\bh} = \bh + \alpha\frac{\bw_{s, c}}{\|\bw_{s, c}\|}$,
%\begin{align}
%    \widehat{\bh} = \bh + \alpha\frac{\bw_{s, c}}{\|\bw_{s, c}\|}
%\end{align}
where $\alpha$ is a scaling factor that controls the strength of the intervention.

\paragraph{TPR probe interventions.}
For TPR probes, the decoding of each square--color pair is now bilinear:
$\ell_{s, c} = \br_{s}^\top\bB\bff_{c}$.
$\bB$ wires square and color vectors such that the bilinear product is high when $s$ is occupied by $c$.
Recall that the binding $\bB$ can be expressed as a sum of outer products, $\sum_{s \in \mathcal{S}} \br_{s} \bff_{y(s)}^\top$
where $y(s)$ denotes the color of square $s$.
One simple intervention is to modify the binding $\bB$ by swapping a single outer product $\br_{s} \bff_{c}^\top$.
To intervene on square $s$ from its original color $y(s)$ to a new color $\widehat{y}(s)$, we replace the outer product $\br_{s} \bff_{y(s)}^\top$ with $\br_{s} \bff_{\widehat{y}(s)}^\top$, after which we map the change in the binding matrix ($\Delta\bB$) back to the hidden state space using the pseudo-inverse of $\bM$:
\begin{alignat}{2}
\label{eq:intervention}
    \Delta\bB &:= \br_{s} (\bff_{\widehat{y}(s)} - \bff_{y(s)})^\top, &\quad&
    \bM_\text{flat} := \operatorname{reshape}(\bM, (d_r d_f) \times d_{model}), \\
    \bz_s &:= \bM_\text{flat}^+ \operatorname{vec}(\Delta\bB), &\quad&
    \widehat{\bh} = \bh + \alpha\frac{\bz_s}{\|\bz_s\|}
\end{alignat}

Interventions with trilinear probes is the same procedure, but replaces $\Delta\bB$ with $\Delta\bT$:
\begin{align}
    \Delta\bT_{ij} &:= \bu_i \otimes \bv_j \otimes (\bff_{\widehat{y}(i,j)} - \bff_{y(i,j)})
    %\bM_\text{flat} &:= \operatorname{reshape}(\bM, (d_{u}d_{v}d_f) \times d_{model})
    %\bz_{ij} &:= \bM_\text{flat}^+ \operatorname{vec}(\Delta\bT_{ij}), &\quad
    %\widehat{\bh} &=
        %\bh + \alpha \frac{\bz_{ij}}{\|\bz_{ij}\|}
\end{align}
\vspace{-10pt}

\para{Composing multiple interventions.}
Since both linear and TPR probe interventions are linear operations, we should be able to compose multiple interventions by simply adding more intervention vectors.
Imagine we wish to alter the color of $k$ squares $\{s_i\}_{i=0}^{k-1}$ from their original colors $\{y(s_i)\}_{i=0}^{k-1}$ to alternate colors $\{\widehat{y}(s_i)\}_{i=0}^{k-1}$.
Interventions can then be composed as follows:

\vspace{-1.5em}
\begin{align}
    \small{
        \text{Linear}: \widehat{\bh} = \bh +
            \sum_{i=0}^{k-1} \alpha_i\frac{\bw_{s_i, \widehat{y}(s_i)}}{\|\bw_{s_i, \widehat{y}(s_i)}\|},
        \quad
        \text{TPR (Bilinear)}: 
            \Delta\bB := \sum_{i=0}^{k-1} \beta_j \br_{s_i} (\bff_{\widehat{y}(s_i)} - \bff_{y(s_i)})^\top,
    }
\end{align}
\vspace{-0.5em}

where $\beta_j$ is also a scaling factor and the remaining steps are the same as the single intervention case.
A similar composition can be done for trilinear probes by replacing $\Delta\bB$ with $\Delta\bT$.
For $\alpha_i, \beta_j$, we sweep over all possible combinations over values $\{0.25, 0.5, 0.75, \dots 2.5\}$ per test sample.
We intervene on every layer at the last timestep.

\begin{wrapfigure}{r}{0.565\textwidth}
\centering
    \includegraphics[width=0.99\linewidth]{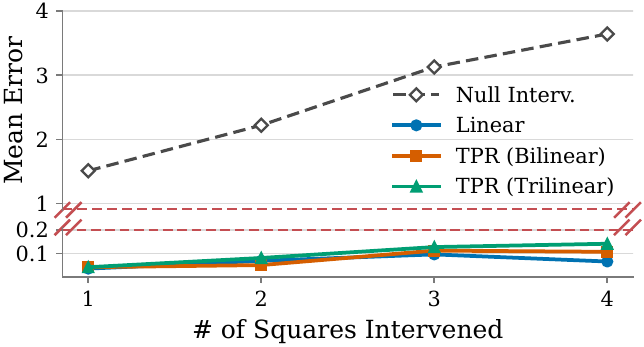}
    \caption{\textbf{Intervention results for linear \& TPR probes.}}
    \label{fig:intervention_results}
\end{wrapfigure}

\para{Evaluation.}
We test our interventions on 1,000 held out samples.
We compare the intervened move predictions $\bmoves_{interv} \in \{0, 1\}^{8 \times 8}$ against the groundtruth set of valid moves $\bmoves_{target} \in \{0, 1\}^{8 \times 8}$ corresponding to the target board-state $\bBoard^{(target)}$.
Per prior work~\citep{li2022emergent, nanda2023emergent}, we report \emph{mean error count} -- the average number of false positives and false negatives given $\bmoves_{interv}$ and $\bmoves_{target}$.
As a baseline measure we report the average number of errors under a null intervention -- i.e., the number of errors if we simply compare the model's original next-move predictions $\bmoves_{orig}$ against $\bmoves_{target}$.

\paragraph{Results.}
Figure~\ref{fig:intervention_results} shows the results.
For our bilinear TPR probe, we use $d_r=52,d_f=2$ and for our trilinear probe we use $d_u,d_v=8,d_f=2$.
All interventions achieve near zero error counts, even when multiple interventions are composed.
This confirms that the TPR probes have not only recovered a valid structure, but also that it captures causal mechanisms for next-move prediction.

\subsection{TPR vs. Linear Directions}
\label{subsec:tpr_vs_linear}

How do our recovered structure relate to linear probes?
We study this question by deriving ``effective linear probes'' $\mathbf{\widetilde{W}}_{s, c}$ from the parameters of our TPR probes, by projecting them onto the activation space, the same space from which linear probes were trained:

\vspace{-1em}
\begin{align}
    \text{Bilinear}: \mathbf{\tilde{w}}_{s,c} := \bM_\text{flat}^\top \mathrm{vec}(\br_{s} \bff_c^\top), % \in \mathbb{R}^{d_{\mathrm{model}}},
    \quad
    \text{Trilinear}: \mathbf{\tilde{w}}_{ij,c} := \bM_\text{flat}^\top \mathrm{vec}(\bu_{i} \otimes \bv_{j} \otimes \bff_c) \in \mathbb{R}^{d_{\text{model}}}.
\label{eq:effective_linear_probe}
\end{align}

\begin{figure}
\centering
\includegraphics[width=0.94\linewidth]{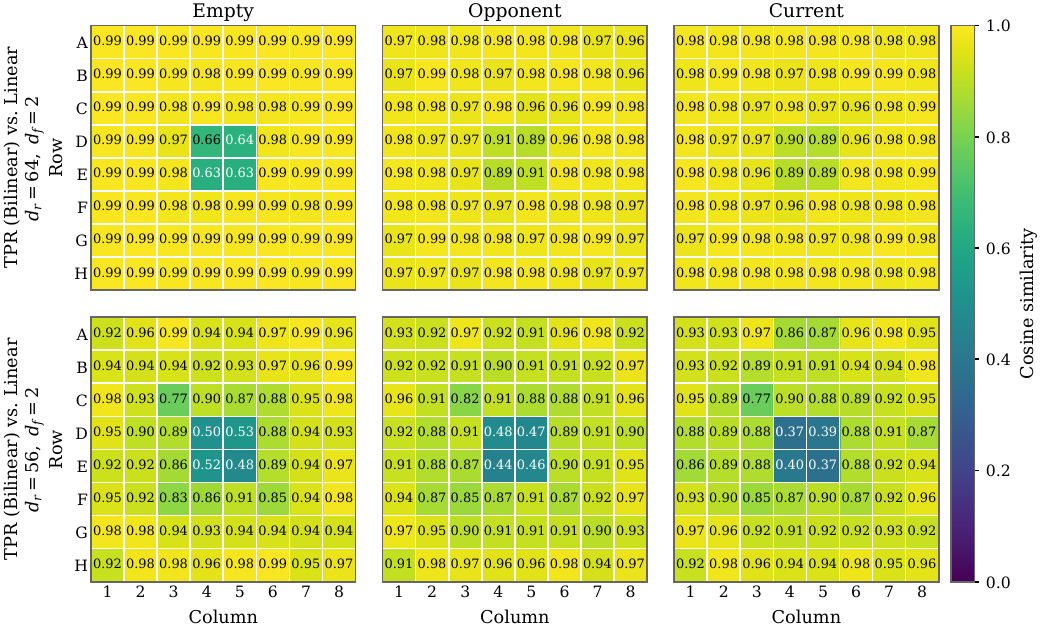}
\caption{\textbf{Cosine similarity scores between linear probes vs. ``effective linear probes'' from TPR probes.}
Linear probes can be recovered from the parameters of our TPR probes, suggesting that linear directions may be a projection of more structured underlying components.
}
\label{fig:tpr_vs_linear_bilinear}
\vspace{-1em}
\end{figure}

We then measure the cosine similarity between these effective linear probes $\mathbf{\widetilde{W}}$ and the independently trained linear probes $\mathbf{W}$.
Because the probes go through a softmax, each (effective) probe could be offset by a constant vector without affecting its logits.
Thus we mean-center $\bW$ and $\widetilde{\bW}$ before taking cosine similarities.
Figure~\ref{fig:tpr_vs_linear_bilinear} shows the results for two bilinear TPR probes: a ``full-dimensional'' one with $d_r=64$ and a ``compressed'' TPR probe with $d_r=56$ (see Figure~\ref{fig:appx_tpr_vs_linear_trilinear} for trilinear probes).

In the full-dimensional case, we see near perfect alignment with the linear probes for all square--color pairs, suggesting that the TPR probe learns a simple reparameterization of $\bW \in \mathbb{R}^{192 \times d_\text{model}}$, which we validate in Appendix~\ref{sec:appx_full_dimensional_tpr_vs_linear}.
In the latter compressed case, the TPR probe no longer has enough dimensions to simply reshape $\bW$, as it must go through the binding matrix $\bB \in \mathbb{R}^{d_r \times d_f}$ as a bottleneck.
However, we still see significantly high cosine similarity scores, suggesting that the same linear directions can be nearly recovered with our TPR probes.

To summarize, the TPR probes learn the same ``effective'' directions as the linear probes, but with additional structure.
Put differently, the linear directions admit a structured factorization, though we refrain from claiming that the model itself natively represents the board in TPR form.
Because the full-dimensional case is simply a reparameterization of the linear probes, for the rest of the analyses we use a compressed TPR probe with $d_r=52$.

\para{Local vs. distributed codes.}
The two cases above are closely related to the coding scheme described in \cite{Thorpe1989LocalVD}.
A local code encodes each concept (e.g., a square--color pair) with a single dimension, whereas a distributed code encodes a concept across multiple dimensions.
When $d_r = 64$, the TPR probe has enough dimensions to index all linear probes, yielding a local code of all 192 square--color probe directions (see Appendix~\ref{sec:appx_full_dimensional_tpr_vs_linear}).
With $d_r < 64$, the probe no longer has enough dimensions, yielding a distributed code (i.e., superposition).

\subsection{Structural Decomposition or Simply Low-Rank?}
\label{subsec:tpr_vs_svd}

\begin{wrapfigure}{r}{0.53\textwidth}
\centering
    \includegraphics[width=0.99\linewidth]{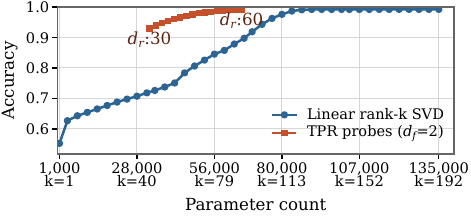}
    \caption{\textbf{Rank-k truncated SVD accuracy.}
    At rank-80, $\texttt{SVD}_k(\bW)$ matches the number of parameters as our TPR probe, but only achieves 85\% accuracy.
    }
    \label{fig:tpr_vs_svd}
    \vspace{-0.5em}
\end{wrapfigure}

Thus far we show that TPR yields a structural low-rank decomposition of the linear probes $\bW$ -- is this because $\bW$ is simply low-rank?

Thus we compare the accuracy of rank-$k$ truncated SVD of $\bW$ (denoted $\texttt{SVD}_k(\bW)$) against our TPR probes ($d_r=30\sim60$).
We sweep over $k$ and compare both the accuracy but also the number of parameters in $\texttt{SVD}_k(\bW)$: see Figure~\ref{fig:tpr_vs_svd}.
The x-axis indicates the number of parameters in $\texttt{SVD}_k(\bW)$ and the y-axis indicates probing accuracy.
Compared to our TPR probe with $d_r=52$, $\texttt{SVD}_k(\bW)$ matches the number of parameters at $k=80$ but only reaches 85\% accuracy, while reaching 99\% accuracy at $k=120$ (150\% of the TPR probe's number of parameters).
This suggests that the TPR probe has learned a structural decomposition beyond just a low-rank decomposition.

\subsection{Geometric Signatures of Board Structure in TPR Weights}
\label{subsec:board_structure}

Figure~\ref{fig:main_figure} suggests that the learned square embeddings $\bR \in \mathbb{R}^{64 \times d_r}$ recover a geometry aligned with the structure of the Othello board.
We now quantify this observation.
Recall that each row $\br_s \in \mathbb{R}^{d_r}$ corresponds to a square $s$ on the board, and we use a TPR probe with $d_r=52$.

\para{Baselines.}
To distinguish the geometry induced by OthelloGPT from that induced by either the TPR architecture or board-state statistics alone, we compare against two baselines.
\textbf{(i) TPR-OOD} is a TPR probe trained on out-of-distribution board-states in which the colors of the squares are sampled independently at random.
These are likely invalid board-states, but allow us to test whether the TPR architecture alone induces board-aligned geometry.
\textbf{(ii) TPR-Random Coding} is a TPR probe trained on the same distribution of board-states as the original TPR probes, but replaces OthelloGPT activations with synthetic encodings.
Namely, for each square--color pair $(s, c)$, we sample a random vector $\bq_{s, c} \in \mathbb{R}^{d_\text{model}}$ and encode each board-state by summing the 64 corresponding square--color vectors.
This baseline tests how much geometry can be explained by the board-state distribution together with the TPR architecture, rather than by structure present in OthelloGPT's activations.
Both baseline probes achieve 99\% accuracy on their respective in-domain distributions.

\begin{wrapfigure}{r}{0.57\textwidth}
\centering
    \includegraphics[width=0.99\linewidth]{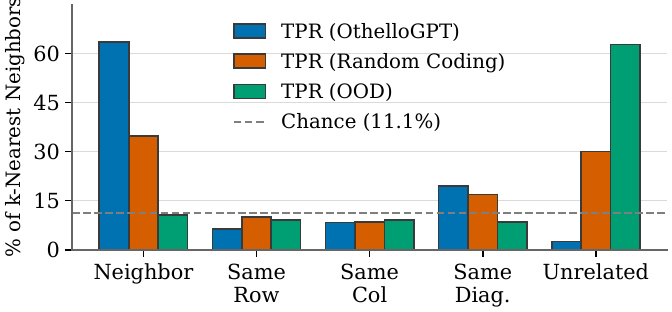}
    \caption{\textbf{Local k-NN based classification of neighbors.}}
    \label{fig:local_geometry_knn}
\end{wrapfigure}

\para{Local neighborhood structure.}
We first evaluate local neighborhood structure.
Depending on its position, each square $s$ has $k_s$ adjacent board neighbors (including diagonals).
For each square $s$, we retrieve the $k_s$-nearest neighbors of $\br_s$ in embedding space and classify each retrieved square into five disjoint categories: true board neighbor, same row, column, diagonal, or unrelated.

As shown in Figure~\ref{fig:local_geometry_knn}, the square embeddings recovered from OthelloGPT exhibit much stronger local agreement with the board than either baseline: roughly 60\% of the retrieved nearest neighbors are true board neighbors, and most of the remaining retrieved squares lie on the same row, column, or diagonal.
In contrast, the baselines retrieve substantially more unrelated squares.
This suggests that the square embeddings recovered from OthelloGPT reflect local board geometry beyond what is induced by the TPR architecture or data distribution alone.

\para{Pairwise board geometry.}
We next evaluate pairwise board geometry.
For each pair of squares $s=(i, j), s'=(i', j')$, we compute their row and column gaps $\Delta i = |i - i'|$ and $\Delta j = |j - j'|$.
We then group all $\binom{64}{2} = 2016$ square pairs by $(\Delta i, \Delta j)$, and compute the average cosine similarity of $\br_s$ and $\br_{s'}$ within each group:
\begin{align}
    \text{GapSim}(\Delta i, \Delta j) = \mathbb{E}_{s, s': |i - i'| = \Delta i, |j - j'| = \Delta j} \left[ \text{cos}(\br_s, \br_{s'}) \right]
\end{align}
\vspace{-1.5em}

\begin{figure}
\centering
\includegraphics[width=0.95\linewidth]{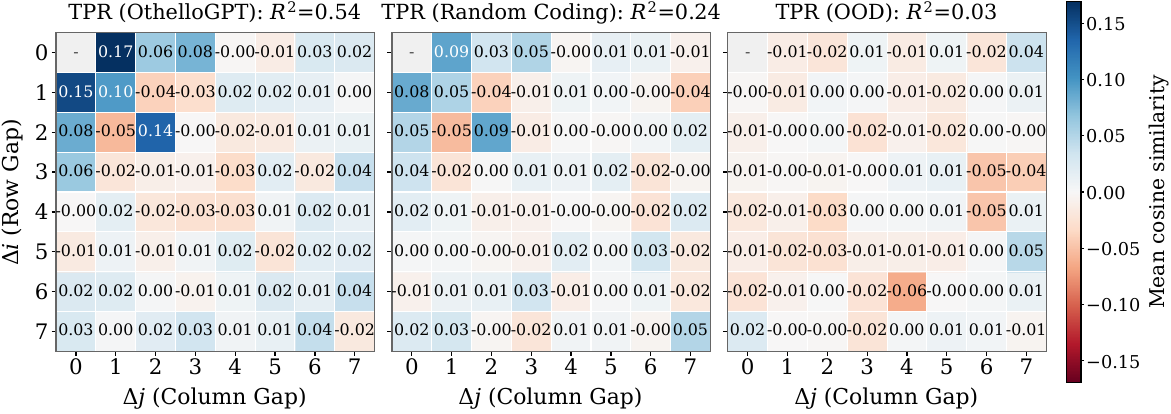}
\caption{\textbf{Pairwise Board Geometry.}
Each entry shows the average cosine similarity between pairs of square embeddings that are $\Delta i$ rows and $\Delta j$ columns apart on the board.
Pairs that are close on the same row ($\Delta i =0$), column ($\Delta j = 0$), or diagonal ($\Delta i = \Delta j$) exhibit higher cosine similarity.
}
\label{fig:local_geometry_heatmap}
\vspace{-1em}
\end{figure}

This gives an $8 \times 8$ matrix of average cosine similarities, where the entry at row $\Delta i$ and column $\Delta j$ corresponds to the average cosine similarity between pairs of squares with row gap $\Delta i$ and column gap $\Delta j$.
Figure~\ref{fig:local_geometry_heatmap} shows that nearby squares on the same row ($\Delta i = 0$), column ($\Delta j = 0$), or diagonal ($\Delta i = \Delta j$) have higher cosine similarity.
Some of this structure also appears in the random-coding baseline, indicating that the TPR architecture and board-state distribution can induce some board-aligned geometry.
However, the effect is strongest for the TPR probe trained on OthelloGPT.

To summarize this effect quantitatively, we evaluate how much of the variance in pairwise cosine similarities is explained by row and column gaps.
For each pair of squares, we use the corresponding gap-based average $\text{GapSim}(\Delta i, \Delta j)$ as the predictions for its cosine similarity.
This yields a $R^2$ score of 0.54 for TPR-OthelloGPT, compared to 0.24 for TPR-Random Coding and 0.03 for TPR-OOD.
The pairwise geometry of the embeddings recovered from OthelloGPT is substantially more aligned with board-relative row and column gaps than the geometry recovered by either baseline.

Overall, these results suggest that the learned square embeddings recover board-aligned geometry that is not explained by the TPR architecture or board-state statistics alone.

\section{Related Work}
\label{sec:related_work}

Here we provide an abridged overview of related work, with a more extensive one in Appendix~\ref{appx:related_work}.

Tensor product representations~\citep{smolensky1990tensor} have long provided a framework for encoding compositional or relational structure in neural networks~\citep{huang2018tensor, binding2018learning, park2024attention, schlag2018learning}.
In Transformers, TPR-style ideas have been used to improve performance on tasks such as mathematics and abstractive summarization~\citep{schlag2019enhancing, jiang-etal-2021-enriching}.
Closest to our setting, \cite{mccoyrnns} show that RNN hidden-states can be reconstructed using Tensor Product Decomposition Networks.

Meanwhile, a growing body of work finds that features are not always well described by isolated rank-1 directions~\citep{mueller2025isolation}.
Rather, researchers are identifying low-dimensional manifolds and coordinate systems for concept representations~\citep{engels2024not, kantamneni2025language, modell2025origins, gurnee2025when, sarfati2026shape, lee2026decomposing}.

This shift has motivated new methods for recovering faithful structure from model activations.
One line of work develops geometry-aware sparse autoencoders (SAEs) based on explicit structural assumptions~\citep{hindupur2025projecting, costa2025flat, bhalla2025temporal, bussmann2025learning}.
%Another line of work instead uses generative models to infer latent structure without committing to a particular geometric form in advance~\citep{luo2026learning}.
Our work highlights a complementary issue, as prior decompositions do not consider the possible \emph{interactions} (i.e., binding) amongst components.
%Our work shows one possibility of recovering such interactions using tensor product representations.

\section{Discussions, Limitations}
\label{sec:discussion}

We study the dichotomy between linear directions and structured representations by studying OthelloGPT, a model with known linear representations yet trained in a domain with inherent structure.
Our TPR probes factorizes shared structure across linear probes that can not only reconstruct the independently trained linear probes, but also exhibit geometry that reflects the structure of Othello.
Could it be that linear directional representations in general are projections of more complex structures hiding underneath?
We conclude with a few thoughts:

\para{Structured Decoding.}
Our work demonstrates that linear directions can be factorized into components with shared structure.
One limitation of TPR probes is that one must know a priori what structure to look for.
While our specific TPR configuration is not meant to be a universal fit for all domains, we anticipate domain-specific structural probes to recover faithful components across various models from different domains.

\para{Unsupervised Structure Recovery.}
While SAEs have become a popular method for decomposition, they typically lack structure by treating concepts as a bag of linear directions.
While other geometry-aware methods have been suggested, they do not account for the possible interactions between components (e.g., binding).
Could it be that some of the latents being recovered by SAEs correspond to such component-wise interactions?
If so, how might we interpret them?
Furthermore, while our TPR probes only have a single layer of binding, one could have multiple layers to represent hierarchical or nested structure.

\para{TPR vs. Feature Subspaces, Mechanisms.}
Note that we are \emph{not} claiming that OthelloGPT performs tensor products, nor that we have recovered structural components that the model uses inherently (e.g., separate square or color subspaces), although Section~\ref{subsec:board_structure} does reveal some related geometric structure.
Rather, we demonstrate that TPRs can provide a factorization of shared structure across a bag of linear directions.
We do not study how our recovered structures relate to OthelloGPT's mechanisms and leave this exploration for future work.

\clearpage

\begin{ack}

AL thanks Andy Arditi, Eric Michaud, Kiho Park, and Or Shafran for useful discussions and feedback.
In particular, AL thanks Eric for suggesting the comparison between TPR probes against rank-k truncated SVD, as well as with helping us make precise our claims, Andy Arditi for the ``Random Coding'' baseline, and Or for also helping us make our claims more precise.
Lastly, AL thanks Harvard FAS RC for GPU compute.
The authors acknowledge support from a Superalignment Fast Grant from OpenAI, and Coefficient Giving.
\end{ack}

\bibliographystyle{plainnat}
\bibliography{references}

%%%%%%%%%%%%%%%%%%%%%%%%%%%%%%%%%%%%%%%%%%%%%%%%%%%%%%%%%%%%

\clearpage
\appendix
\section{Related Work}
\label{appx:related_work}

\begin{figure}
\centering
    \includegraphics[width=0.99\linewidth, trim={5cm, 3cm, 5cm, 3.5cm}, clip]{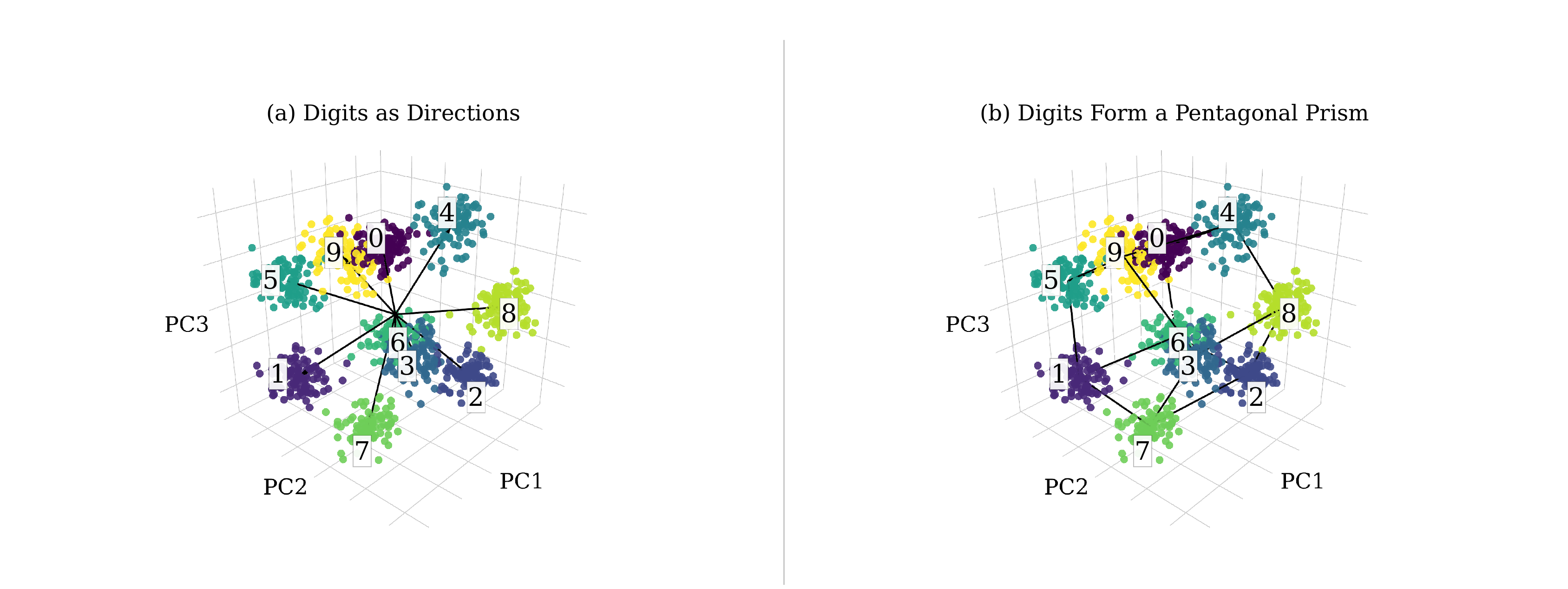}
\caption{Representations of digits in a Transformer trained on multi-digit multiplication may appear as linear directions, but a closer look reveals structure in the form of a pentagonal prism~\citep{bai2025can}.
Similarly, linear directions in language models may be encoding underlying structure.
}
\label{fig:linear_vs_fourier}
\vspace{-1em}
\end{figure}

\paragraph{Tensor product representations.}
Tensor product representations~\citep{smolensky1990tensor} have influenced neural networks in numerous ways, often aimed at representing compositional or relational structure~\citep{huang2018tensor, binding2018learning, schlag2018learning, park2024attention}.
In the context of Transformers, TPRs-inspired architectures have improved performance on structured tasks such as mathematical tasks~\citep{schlag2019enhancing} or abstractive summarization~\citep{jiang-etal-2021-enriching}.
Perhaps closest in spirit to our work, \cite{mccoyrnns} show that RNN hidden-states can be reconstructed using Tensor Product Decomposition Networks.

\paragraph{Feature geometry.}
In recent years, a large body of interpretability work have found numerous concepts that are encoded as linear directions, and that these representations often generalize across models~\citep{lee2025shared}.
Examples include sentiment~\citep{tigges2023linear}, toxicity~\citep{lee2024mechanistic}, refusal~\citep{arditi2024refusal}, ``correctness''~\citep{lee2025geometry}, and even user-attributes~\citep{chen2024designing}.

A growing line of work suggest that many features are not best described with rank-1 linear directions, but instead occupy low-dimensional manifolds or coordinates in a subspace.
Examples include circular geometry for periodic concepts like days of the week~\citep{engels2024not}, but also emotions~\citep{sun2026valence}, helical structure for number representations~\citep{kantamneni2025language}, and manifold structure for dates and years~\cite{modell2025origins}.
Others find features represented as coordinates in low-rank subspaces: \cite{park2024iclr} find that models can represent in-context learning tasks, while \cite{lee2026decomposing} identify low-rank feature interactions in attention mechanisms.
\cite{gurnee2025when} similarly show that features such as word count and token position can lie on manifolds that are aligned to produce high attention scores.

Taken together, these works suggest that linear probes may sometimes only recover local readouts of a richer underlying structure.
A good example might be of \cite{bai2025can}, who study a toy Transformer trained on multi-digit multiplication: see Figure~\ref{fig:linear_vs_fourier}.
The model forms clusters to represent each digit, from which linear directions (i.e., a vector towards the centroids of each cluster) can decode a predicted digit from hidden-states.
However, the clusters themselves form a highly intuitive structure by organizing into a pentagonal prism that reflects parity (even versus odd) and modulo-5 relationships.
This example illustrates the broader possibility that linear directions may be projections of more structured, domain-specific representations.

Perhaps most relevant to our work are three recent lines of work.
\cite{shai2026transformers} find that Transformers can learn factorized representations of underlying latent variables of data generating processes.
\cite{yocum2025neural} argue that neural networks may encode structured feature fields, which linear probes can recover from their activations.
\cite{sarfati2026shape} build on this idea to recover a manifold of ``posterior beliefs'': by training a family of linear probes across different latent parameter settings of a controlled in-context learning task, and by ``tiling'' the linear probes together, they are able to recover a manifold over inferred latent parameter values, similarly suggesting that linear readouts stem from underlying structure.

\paragraph{Unsupervised structure discovery.}
Another related line of work asks how to uncover latent, unknown structure from activations at scale.
Various geometry-aware sparse autoencoders have been proposed, based on various structural assumptions~\citep{hindupur2025projecting, costa2025flat, bhalla2025temporal, bussmann2025learning, muchane2025incorporating}.
A complementary approach is to make no assumptions regarding the underlying states and to rely on generative models to infer the latent structures from the activations~\citep{luo2026learning}.

We envision future steps of structure discovery to be a mix of supervised, domain or concept-specific architectures such as our TPR probe as well as scalable, unsupervised, unrestrictive architectures to recover unknown structure.

\section{Additional Results}
\label{appx:additional_results}

\begin{figure}
\centering
\includegraphics[width=0.95\linewidth]{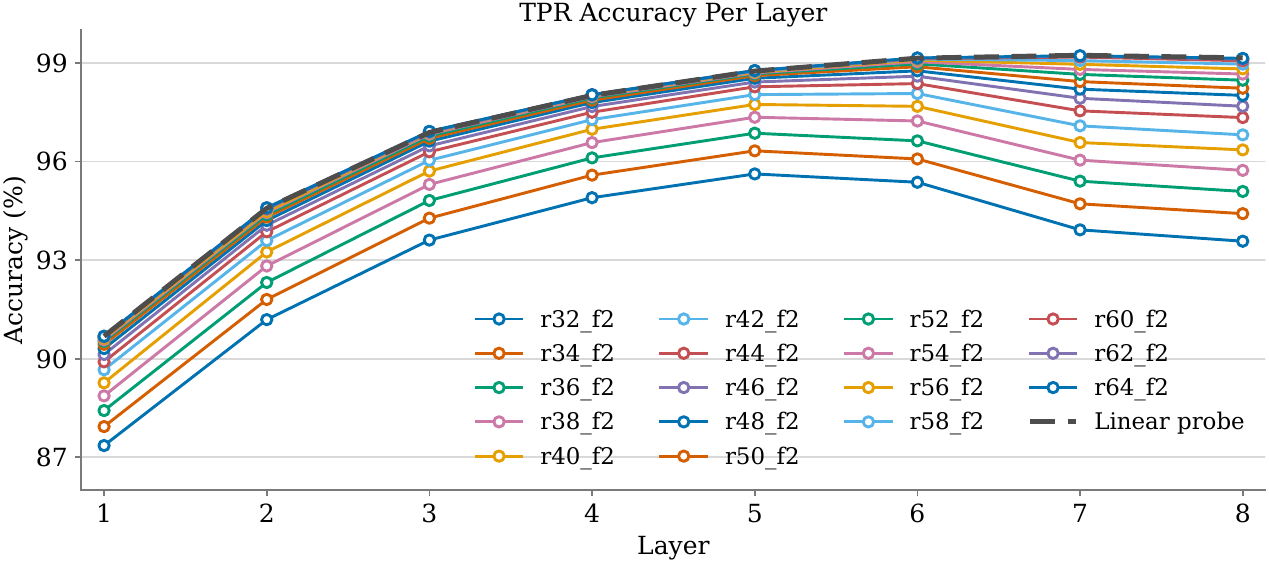}
\caption{\textbf{TPR probe accuracy per layer.}
}
\label{fig:appx_tpr_accuracy_per_layer}
\end{figure}

Figure~\ref{fig:appx_tpr_accuracy_per_layer} shows the TPR probe accuracy per layer for a range of $d_r$ values.
Note that with enough dimensions, the TPR probe achieves the same accuracy as the linear probes, because the TPR probe effectively learns the same directions as the linear probes (see Section~\ref{subsec:tpr_vs_linear}).

Figure~\ref{fig:appx_tpr_vs_linear_trilinear} shows the analogous of Figure~\ref{fig:tpr_vs_linear_bilinear} but for the trilinear TPR probe instead of the bilinear.

%Figure~\ref{fig:appx_simplex_col} shows the analogous of Figure~\ref{fig:simplex_row} but for the column embeddings $\bV$ instead of the row embeddings $\bU$.

%\begin{figure}
%\centering
%\includegraphics[width=0.99\linewidth]{figures/simplex_col.pdf}
%\caption{\textbf{Gram matrices of column-embedding vectors.}
%    This figure is the analogous of Figure~\ref{fig:simplex_row} but for the column embeddings $\bV$ instead of the row embeddings $\bU$.
%    Given column-embeddings $\bV \in \mathbb{R}^{8 \times d_v}$, with enough dimensions ($d_v = 8$) the Gram matrix $\bV\bV^\top$ shows that $\bV$ forms an orthogonal set of basis vectors to encode each column of the board.
%    The mean-centered Gram matrix reveals that $\bV$ further forms a (near) regular simplex.
%    With fewer dimensions ($d_v < 8$), we see a distributed encoding (i.e., superposition) of the 8 columns of the board.
%}
%    \label{fig:appx_simplex_col}
%\end{figure}

%\input{appendix/simplex}
\section{Full-Dimensional TPR Reparameterizes Linear Probes}
\label{sec:appx_full_dimensional_tpr_vs_linear}

\begin{figure}
\centering
\includegraphics[width=0.99\linewidth]{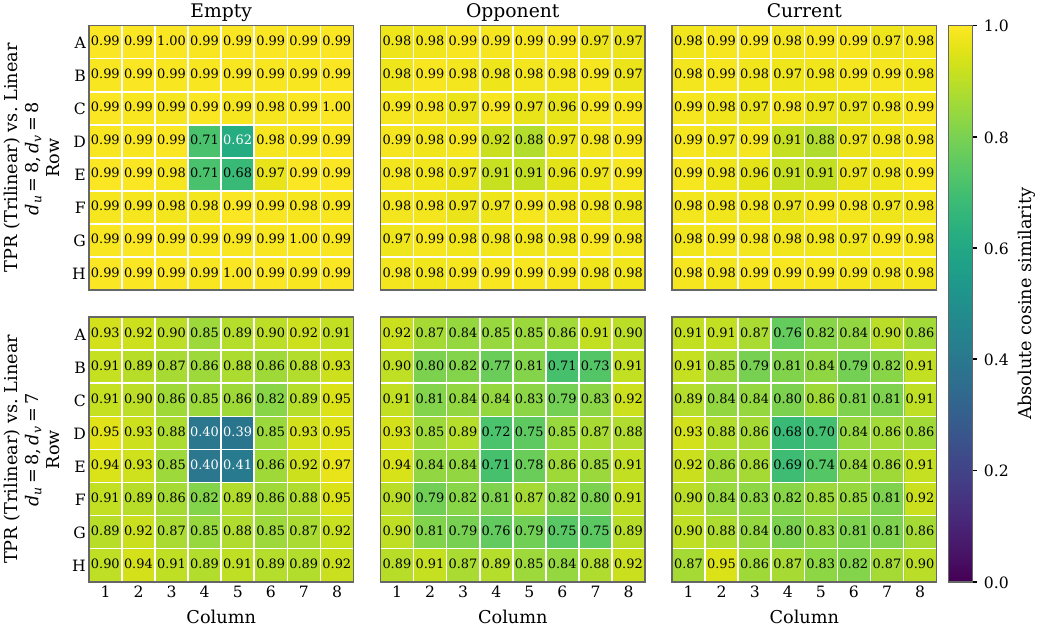}
\caption{\textbf{Cosine similarity scores between linear probes vs. ``effective linear probes'' from trilinear TPR probes.}
}
\label{fig:appx_tpr_vs_linear_trilinear}
\end{figure}

In Section~\ref{subsec:tpr_vs_linear} we see that TPR probes with enough dimensions recover ``effective linear directions'' that closely align with independently trained linear probes.
This occurs for the trilinear TPR probe as well (Figure~\ref{fig:appx_tpr_vs_linear_trilinear}), for a model with $d_u=d_v=8, d_f=2$ dimensions.
Here we explain why this occurs in the full-dimensional setting.

Consider first the trilinear TPR probe,
\[
\bT := \bM(\bh) \in \mathbb{R}^{d_u \times d_v \times d_f},
\qquad
\ell_{ij,c} = \langle \bT, \bu_i \otimes \bv_j \otimes \bff_c \rangle ,
\]
where $\bu_i \in \mathbb{R}^{d_u}$ and $\bv_j \in \mathbb{R}^{d_v}$ are the row and column embeddings, and $\bff_c \in \mathbb{R}^{d_f}$ is the filler embedding for color $c$.
The corresponding effective linear probe direction is
\[
\widetilde{w}_{ij,c}
=
\bM_{\mathrm{flat}}^\top
\mathrm{vec}(\bu_i \otimes \bv_j \otimes \bff_c)
\in \mathbb{R}^{d_{\text{model}}}.
\]

When $d_u=d_v=8$, the row and column embeddings have enough capacity to index all rows and columns of the board independently.
In the simplest case, $\bU$ and $\bV$ could be the standard basis matrices.
Then $\bu_i \otimes \bv_j$ selects the $(i,j)$-th slice of the binding tensor, so each board square is assigned its own $d_f$-dimensional latent vector.

Thus $d_u=d_v=8, d_f=2$ is sufficient to represent 64 three-way classifiers i.e., the 192 linear probe directions.
Although each square has three possible labels, $\mathcal{C}=\{\textsc{Empty}, \textsc{Current}, \textsc{Opponent}\}$, our TPR probes use $d_f=2$.
This is sufficient because a three-way softmax has only two identifiable degrees of freedom.
Adding the same scalar offset to all three logits does not change the predicted probabilities, thus for each square, the probe only needs to represent two independent directions to represent three classes.
This idea is analogous to representing $K$ classes by a $(K-1)$-dimensional simplex.
In the learned filler embeddings, PCA reveals exactly this structure: one axis separates \textsc{EMPTY} from occupied squares, while the other separates \textsc{CURRENT} from \textsc{OPPONENT}.
To summarize, a trilinear probe with $d_u=d_v=8, d_f=2$ can represent the same 64 three-way classifiers (linear probes), and the same argument can be applied for bilinear TPR probes with $d_r=64, d_f=2$.

One thing worth noting is that TPR factorizations are not unique.
Transformations applied to one component (row, column, or filler embeddings) can be compensated by another ($\bM$) while leaving the resulting logits unchanged.
For intuition, suppose the row and column embeddings adopt canonical standard basis vectors.

\begin{figure}
\centering
\includegraphics[width=0.99\linewidth]{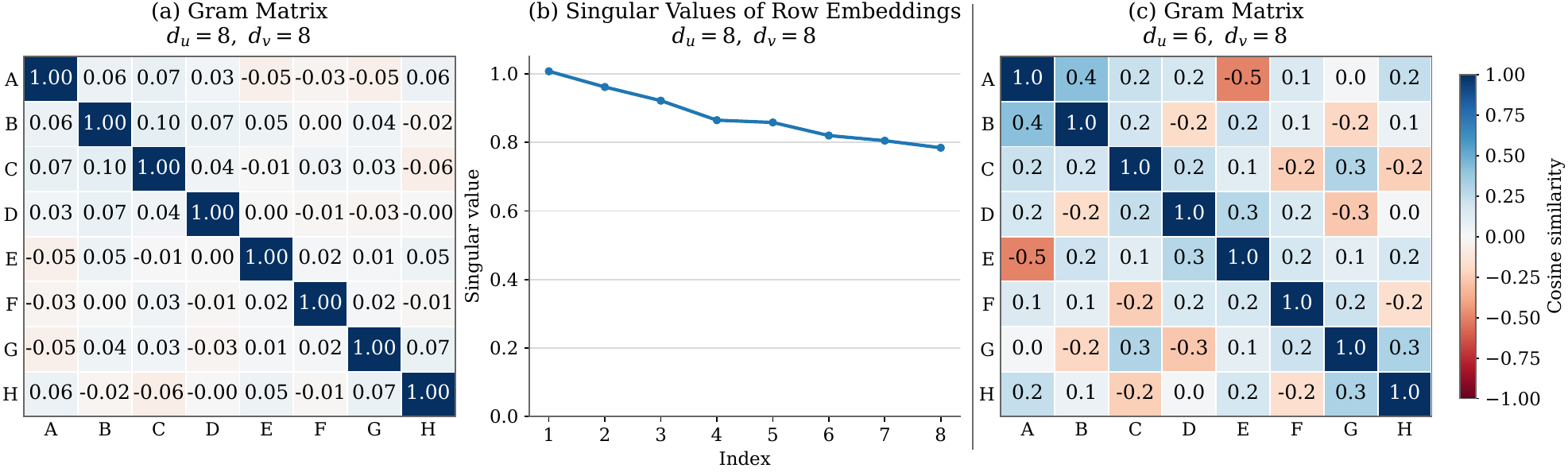}
\caption{\textbf{In a ``full-dimensional'' case, row embeddings effectively behave like an effective orthonormal basis.}
Given row-embeddings $\bU \in \mathbb{R}^{8 \times d_u}$, with enough dimensions ($d_u = 8$) the row-normalized Gram matrix $\bU\bU^\top$ shows that the rows of $\bU$ form an orthogonal set of basis vectors to encode each row of the board.
The singular values of $\bU$ are all close to 1, confirming that $\bU$ behaves like an effective orthonormal basis.
With fewer dimensions ($d_u < 8$), we instead observe a distributed encoding (i.e., superposition) of the 8 rows of the board.
}
    \label{fig:appx_gram_row}
\vspace{-5pt}
\end{figure}

In fact, we empirically observe that row and column embeddings form ``effective'' standard basis vectors.
In the full-dimensional trilinear probe, the recovered row embeddings form an approximately orthogonal basis. To quantify this, we row-normalize $\bU$ and compute the Gram matrix
$\bG_{\bU} = \widetilde{\bU}\widetilde{\bU}^{\top} \in \mathbb{R}^{8\times 8}$,
where each entry gives the cosine similarity between a pair of row embeddings.
Figure~\ref{fig:appx_gram_row}(a) shows that $\bG_{\bU}$ is close to the identity matrix, and Figure~\ref{fig:appx_gram_row}(b) shows that the singular values of $\bU$ are all close to one.
Together, these indicate that the learned row embeddings behave like an effective orthonormal basis.

\begin{figure}
\centering
\includegraphics[width=0.99\linewidth]{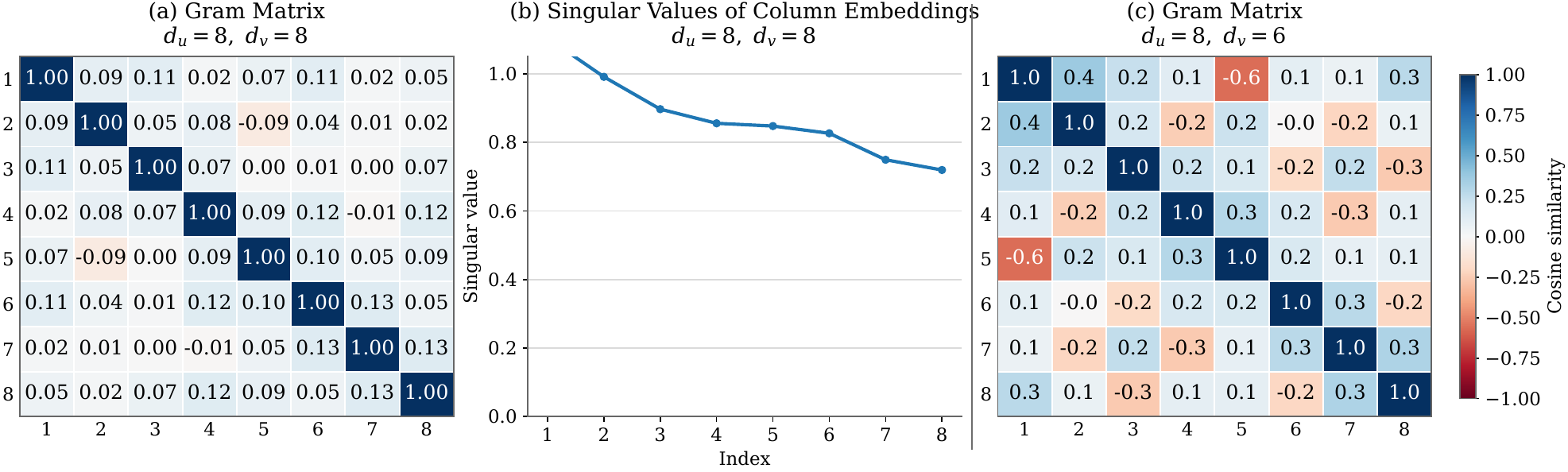}
\caption{\textbf{Gram matrix and singular values of column embeddings.}
}
    \label{fig:appx_gram_col}
\end{figure}
Note that this behavior does not occur with incomplete dimensions.
When $d_u<8$, the probe can no longer assign mutually orthogonal basis vectors to all eight rows.
Instead, the row identities must be represented in superposition.
Figure~\ref{fig:appx_gram_row}(c) illustrates the row-normalized Gram matrix for this setting, in which we no longer observe the identity matrix.

We observe the same patterns for column embeddings in Figure~\ref{fig:appx_gram_col}.

In summary, in the full-dimensional regime, TPR probes can recover the same effective linear directions as standard linear probes because they can allocate independent basis elements to each board position, essentially reparameterizing the linear probes.
In incomplete-dimensional regimes, the TPR probe must compress the board-state representation through a structured bottleneck, leading to a structural distributed code.
\section{Training Details}
\label{appx:training_details}

Table~\ref{appx_tab:hyperparams} provides our hyperparameters for training linear and TPR probes.
All of our experiments are conducted on a single Nvidia H100 80GB GPU, but significantly less memory (16GB) will likely suffice.

\begin{table}
\centering
\caption{Hyperparameters for probes.}
\label{appx_tab:hyperparams}
\begin{tabular}{cc}
\toprule
Learning rate   & 1e-2 \\
Weight decay    & 1e-2 \\
Batch size      & 128 \\
Num epochs      & 1 \\
Validation patience & 10 \\
Validate every & 100 \\
Train size & 295,699 \\
Validation size & 512 \\
Test size & 1,000 \\
\bottomrule
\end{tabular}
\end{table}

\section{Societal Impact}
\label{appx:societal_impact}

Our work takes a step towards better understanding and interpreting the internal representations of language models.
We hope a better understanding will lead to safer and more reliable use cases of models in the future.

%%%%%%%%%%%%%%%%%%%%%%%%%%%%%%%%%%%%%%%%%%%%%%%%%%%%%%%%%%%%

%\newpage
%\input{checklist.tex}

\end{document}